\documentclass[runningheads]{llncs}

\usepackage[final,year=2024,ID=*****]{accv}

\usepackage{accvabbrv}

\usepackage{colortbl}
\usepackage{graphicx}
\usepackage{booktabs}
\usepackage{color}

\usepackage[accsupp]{axessibility}  

\usepackage[pagebackref,breaklinks,colorlinks,citecolor=accvblue]{hyperref}

\usepackage{orcidlink}

\begin{document}

\title{FAGhead: Fully Animate Gaussian Head from Monocular Videos} 


\author{Yixin Xuan\inst{1} \and Xinyang Li\inst{1} \and Gongxin Yao\inst{1} \and Shiwei Zhou\inst{2} \and Donghui Sun\inst{2} \and Xiaoxin Chen\inst{2} \and Yu Pan\inst{1}\thanks{Corresponding Author. Email: ypan@zju.edu.cn}}

\authorrunning{Y. Xuan, X. Li, G. Yao, S. Zhou, D. Sun, X. Chen, Y. Pan et al.}

\institute{Zhejiang University \and vivo AI Lab}

\maketitle

\begin{abstract}

High-fidelity reconstruction of 3D human avatars has a wild application in visual reality. In this paper, we introduce FAGhead, a method that enables fully controllable human portraits from monocular videos. We explicit the traditional 3D morphable meshes (3DMM) and optimize the neutral 3D Gaussians to reconstruct with complex expressions. Furthermore, we employ a novel Point-based Learnable Representation Field (PLRF) with learnable Gaussian point positions to enhance reconstruction performance. Meanwhile, to effectively manage the edges of avatars, we introduced the alpha rendering to supervise the alpha value of each pixel. Extensive experimental results on the open-source datasets and our capturing datasets demonstrate that our approach is able to generate high-fidelity 3D head avatars and fully control the expression and pose of the virtual avatars, which is outperforming than existing works.
\end{abstract}

\begin{figure*}[htb]
\centering
\includegraphics[width=12.3cm]{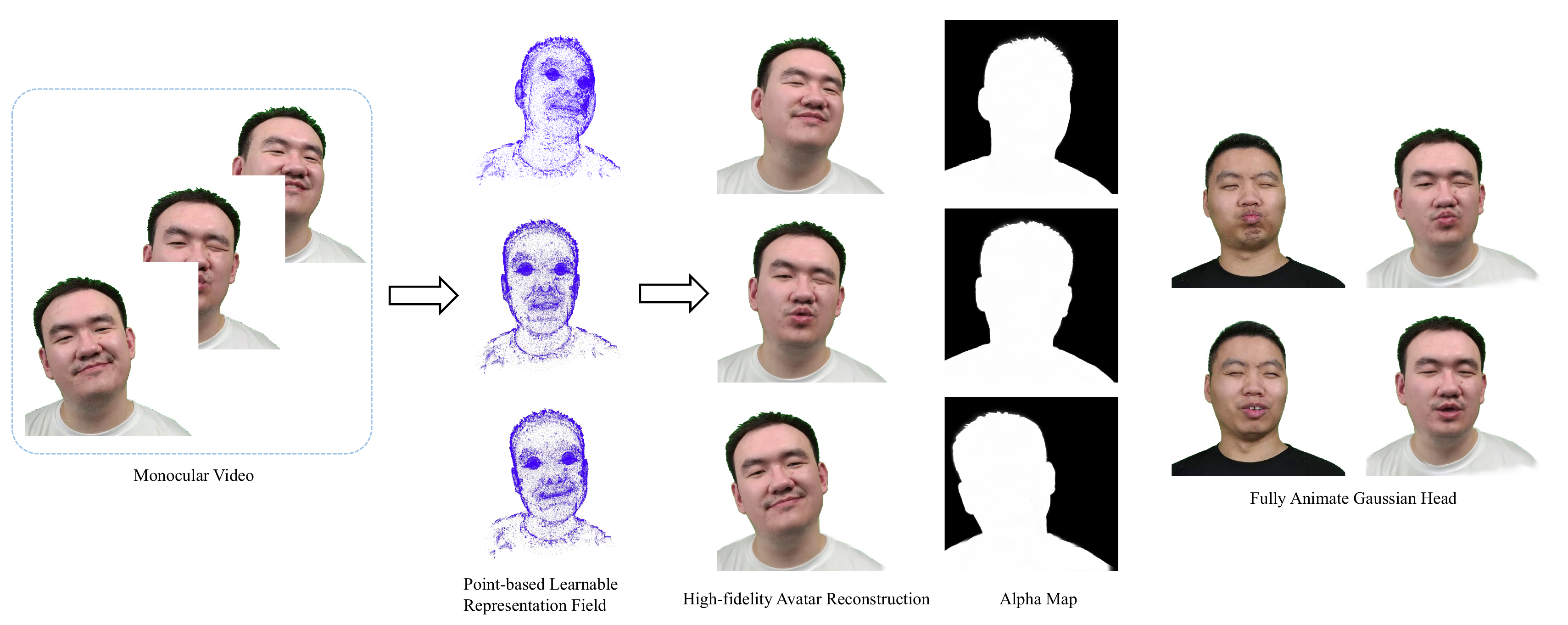}
\caption{
Given the monocular video, our proposed FAGhead approach is able to generate high-fidelity avatars and the corresponding alpha map. By leveraging the novel Point-based Learnable Representation Field, FAGhead ensures photorealistic reanimation and extends generalization to novel expressions and head poses.}
\label{fig:ex0}
\end{figure*}

\section{Introduction}
3D head avatars reconstruction from monocular video has witnessed a significant surge in recent decades, driven by a host of applications such as 3D content creation \cite{chaudhuri2013attribit}, virtual reality(VR) technology \cite{wang2020vr} and gaming \cite{zackariasson2012video}, which is a challenge in the computer vision field. With the development of digital human, the demand for automated photo-realistic avatars synthesis has become more and more prevalent. 

The previous works mainly exploit the 3D morphable models(3DMMs) \cite{blanz2023morphable, paysan20093d} representation, focusing on the shape and expression transform \cite{li2017learning, zielonka2022towards, feng2021learning} to match the original avatars. However, in mono-view settings, these method fail to meet photorealistic requirements and require accurate geometry meshes as priors, which limits their applications.

The advancements in the field of geometry reconstruction have significantly enhanced the accuracy of geometry synthesis. Neural Radiance Field(NeRF) \cite{mildenhall2021nerf} is the most representative work in this field, showing the great capability with complex objects and leading to more high-quality result. Some method \cite{park2021nerfies, park2021hypernerf} produce the photo-realistically human avatars by optimizing an additional continuous volumetric deformation field, while other method \cite{athar2022rignerf, gao2022reconstructing, li2023efficient} combine with the traditional 3DMMs approach and have the capacity to generalize to novel deformations \cite{grassal2022neural}. However, the volume rendering approach, which relies on extensive sampling and alpha compositing, constrain the speed of inference.

Thus, the recent 3D Gaussian Splatting(3DGS) \cite{kerbl20233d} utilized a set of 3D Gaussian points to describe 3D real-world scene, assigning the 3D Gaussian points with variable proprieties, demonstrated the feasibility of photo-realistic novel view synthesis and high efficiency. As for the animated avatars synthesis, maintain approaches \cite{xu2023gaussian, wang2023gaussianhead, chen2023monogaussianavatar, rivero2024rig3dgs} are creating a deform field from canonical to deformation space with the use of a Multi-Layer Perceptron(MLP). Although these approaches have made a profound advance in photo-realistic avatars synthesis, they are unable to decouple identity and expression information effectively, which lead to unreasonable results when facing animation tasks with novel expression.

To overcome this issue and further improve the animation quality, we propose the FAGhead, a novel method based on 3DMMs representation for high-fidelity avatars construction and animation. In spirited by previous works \cite{kirschstein2023nersemble, qian2023gaussianavatars}, which fully explicit the FLAME \cite{li2017learning} model via linear blend skinning (LBS) in multi-view camera setting, we expand it in monocular setting. Regarding decoupling, we separate identity and expression information during preprocessing through a modified face tracker \cite{zielonka2022towards}. 

Regarding Gaussian initialization, we propose a novel Point-based Learnable Representation Field (PLRF) approach that positions Gaussian points along the midline of a single triangle face, thereby increasing the density of Gaussian points and enhancing facial expression details. Specifically, instead of initializing the 3D Gaussian points at the center of each triangle face of the FLAME mesh, we sample Gaussian points with learnable positions along the line segments connecting the centroid to each vertex of the triangle face within each avatar mesh. 

Moreover, a transform network is built to match the dynamic facial movements from canonical point-based field to the transfrom point-bese field. In practice, it takes pre-retrieved FLAME parameters
as conditions to produce the facial movement deformation.
Besides, to enhance the rendering edge performance as hair and shoulder, we introduce the alpha loss between alpha map and the rendered. With the assistance of these enhancements, FAGhead aachieves higher-fidelity rendering and provides fully controllable avatars over facial expressions and head poses.  In summary, our contributions are as follows:

\begin{itemize}
\item[$\bullet$] We propose Fully Animate Gaussian Head, a novel head avatar synthesis approach with effective representation field, which could provide fully controllable head avatars and achieve high-fidelity face reenactment.

\item[$\bullet$] We propose a Point-based Learnable Representation Field(PLRF), which significantly enhance the generated quality and geometry detail structure. 

\item[$\bullet$] We introduce the transform network to fit the deformation defined by 3DMM model. 

\item[$\bullet$] We redesign the model structure and propose the additional loss item as alpha loss item and other regularization loss to enhance the performance of geometry reconstruction. 

\end{itemize}
\section{Related Work}
\subsection{Scene Reconstruction and Novel View Synthesis}
Early novel view synthesis mainly focuses on image interpolation and image-based rendering(IBR) \cite{shum2000review, zhang2004survey}, which achieves novel view synthesis by image interpolation between nearest-neighbour images. In the recent works, Significant advancements have been made in the fields of Structure from Motion (SfM) \cite{schonberger2016structure} and Multi-view Stereo (MVS) \cite{tomasi1992shape}, both of which utilize explicit 3D scene representations through RGB images \cite{buehler2023unstructured, snavely2006photo, han2019image}. Besides, COLMAP \cite{schonberger2016structure} plays a significant role, which provides the reliable 3D space feature points matched across multi-view images. With development of neural rendering \cite{tewari2022advances}, more effective and realistic approaches are proposed, especially Neural radiance fields (NeRF) \cite{mildenhall2021nerf}. NeRF achieves high realistic reconstruction with manipulating an MLP network as 3D space expression and volume rendering. 

To enhance the efficiency of NeRF, Point-NeRF \cite{xu2022point} combines a neural point cloud initialized via a deep network, accelerating training time via a novel point cloud pruning and growing mechanism. NSVF \cite{liu2020neural} uses a sparse voxel octree to represent the scene which avoid the computational waste. Meanwhile, a various of advanced NeRFs employing another explicit representations \cite{fridovich2022plenoxels, yu2021plenoctrees, sun2022direct, chen2022tensorf, jin2023tensoir} are proposed to overcome slow speed issue. However, there is still a host of room for improvement, the render speed cannot meet the real-time requirement. 3D Gaussian Splatting \cite{kerbl20233d} provides a faster and more efficient method for reconstructing the 3D scene, exhibiting a easy and effective performance with high quality.

\subsection{3D Parameter Head Model}
Building on the premise that the human head shape space can be effectively separated into identity, expression, and appearance components. The 3D Morphable Model (3DMM) \cite{blanz2023morphable} is proposed, which is used to embed 3D head shapes into multiple low-dimensional principal component analysis (PCA) spaces. 
Subsequent research \cite{cao2013facewarehouse, tran2018nonlinear, romdhani20053d, hu2004improved} has expanded this mesh-based parametric head model, enhancing its representational capacity through the development of multi-linear \cite{cao2013facewarehouse} and non-linear models \cite{tran2019towards}, as well as articulated models equipped with corrective blendshapes.

Recent cutting-edge techniques have advanced in accurately capturing the intricate deformations associated with facial expressions by incorporating additional displacement maps that respond to the input images \cite{danvevcek2022emoca, feng2021learning}. In addition, generative models powered by machine learning, such as GANs \cite{karras2017progressive} and StyleGAN \cite{karras2019style}, have been integrated into current frameworks \cite{gecer2021fast, luo2021normalized} to refine the precision in modeling facial textures and geometries. However, despite these advances, the parametric models are still mostly limited to capturing the facial region's geometry and appearance at a rather basic level through explicit mesh models. This limitation detracts from the photorealistic quality of reconstructions and animations based on these models \cite{grassal2022neural}. 

\subsection{3D Head Portrait Synthesis}
3D head portrait synthesis could be divide into explicit and implicit representations. The explicit representations, primarily based on mesh models \cite{cao2013facewarehouse}, which have been evolving for many years. In recent efforts, some researchers \cite{kim2018deep, thies2020neural} have utilized 2D neural rendering techniques for creating photo-realistic portraits, although these method often overlook non-facial areas or face challenges with temporal and spatial inconsistencies due to their weak integration with 3D geometry. Other approaches \cite{feng2021learning, grassal2022neural} have focused on learning vertex offsets to capture the detailed head geometry more accurately, but they can still encounter issues with geometry and texture artifacts in complex areas like hair, eyes, and mouth, limited by the mesh model's representational capacity and the challenges of differentiable rendering. PointAvatar \cite{zheng2023pointavatar} introduced a novel, deformable point-based approach, overcoming some mesh model limitations, albeit at the cost of requiring an excessive number of points and extensive training periods. Implicit models, on the other hand, utilize neural functions to create digital head avatars, with significant research \cite{zheng2022imface} dedicated to achieving high fidelity, though often at the expense of training and inference efficiency. Innovations such as volumetric primitives \cite{lombardi2021mixture} and local feature grids \cite{gao2022reconstructing, xu2023avatarmav, zielonka2023instant} have been proposed to enhance efficiency and reduce the computational load. Moreover, FlashAvatar \cite{xiang2023flashavatar} proposed the UV sample strategy to enhance rendering efficiency. 

\section{Preliminary}
There is a enormous difference compare 3D Gaussian Splatting \cite{kerbl20233d} with the widely adopted Neural Radiance Field. 3DGS utilizes explicit 3D Gaussian points which initializes with the feature points generated by COLMAP \cite{schonberger2016structure} as the fundamental entities for rendering. Consider the rendering entities set $\{G_t\}^N_{t=0}$, $t$ is index of 3D Gaussian point, $N$ is the number of 3D Gaussian point. Each point has following proprieties $\zeta_t=\{\mu_t, o_t, r_t, s_t, c_t\}$. $\mu_t\in\mathbb{R}^3$ is the position of Gaussian point, $o_t \in \mathbb{R}$ describes the opacity, $r_t\in\mathbb{R}^4$ indicates the rotation, $s_t\in\mathbb{R}^3$ reflects the scale of Gaussian point, $c_t\in\mathbb{R}^3$ denotes the view-dependent color, calculated via spherical harmonic coefficients, which $k$ is related with the degree of spherical harmonic. Each 3D Gaussian point is mathematically defined with a spatial mean $\mu$ and a covariance matrix $\Sigma$ as following: 
\begin{equation}
\label{eq1}
    G_t(x) = \exp({-\frac{1}{2}(x-\mu)^{T}\Sigma^{-1}(x-\mu)}),
\end{equation}

During rendering, all 3D Gaussian points are projected to the specified 2D camera plane at first. According to the prior work \cite{zwicker2001ewa}, given a viewing transformation $W$, the 3D covariance matrix $\Sigma'$ can be reasoned as follow:
\begin{equation}
\label{eq2}
    \Sigma'=JW\Sigma W^T J^T,
\end{equation}
where $J$ denotes the Jacobian of the affine approximation of perspective projection transformation. 

Thus, the color $\hat{C}$ of specified pixel can be synthesized as:
\begin{equation}
\label{eq3}
    \hat{C}(x) = \sum_{t \in M} c_{t} \alpha_{t}(x) \prod_{j=1}^{t-1} (1 - \alpha_{j}(x))
\end{equation}

with
\begin{equation}
\label{eq4}
    \alpha_t(x) = o_t\exp({-\frac{1}{2}(x-\mu)^{T}\Sigma^{-1}(x-\mu)})
\end{equation}

To optimize shared memory usage, 3DGS has developed a GPU-optimized rasterization process that assigns each thread block to an image tile. This innovative approach not only allows for realistic scene reconstruction but also delivers considerably faster rendering speeds and decreases memory consumption during training, outperforming NeRF approach.

\begin{figure*}[htb]
\includegraphics[width=12.6cm]{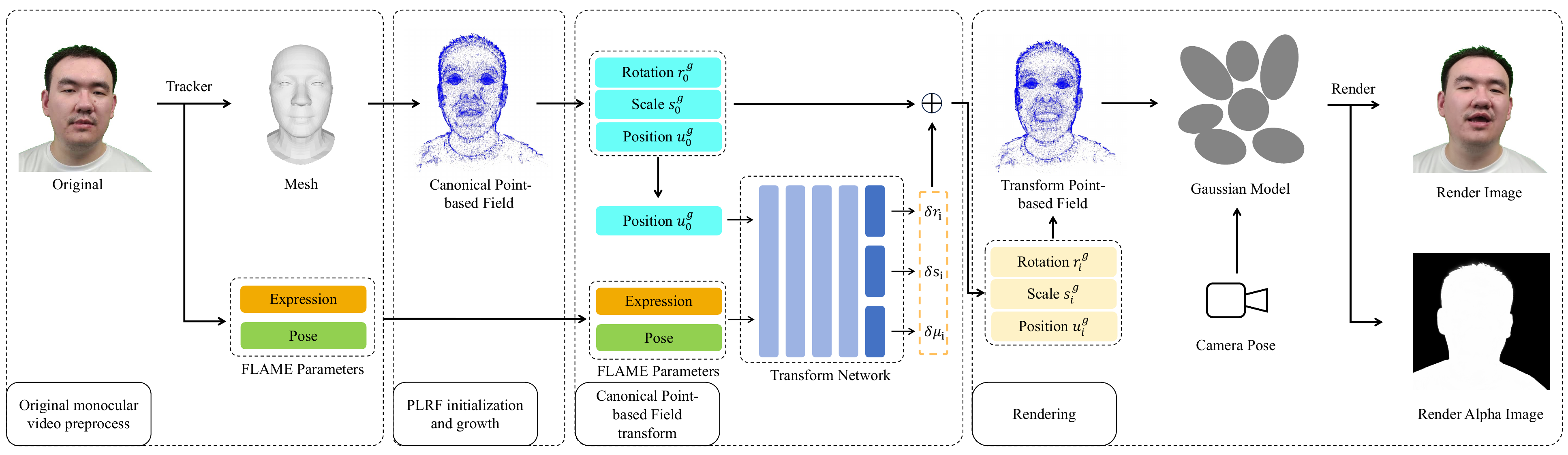}
\caption{
FAGhead overview. before training, a Point-based Learnable Representation Field is established on the timestep 0 as canonical point-based field. Via the transform network that input the canonical global position and current FLAME parameters and output the deformation between canonical and transform space. Besides, we introduce the alpha rendering in order to eliminate the geometry mistake.}
\label{fig:ex1}
\end{figure*}

\begin{figure*}[htb]
\includegraphics[width=11.6cm]{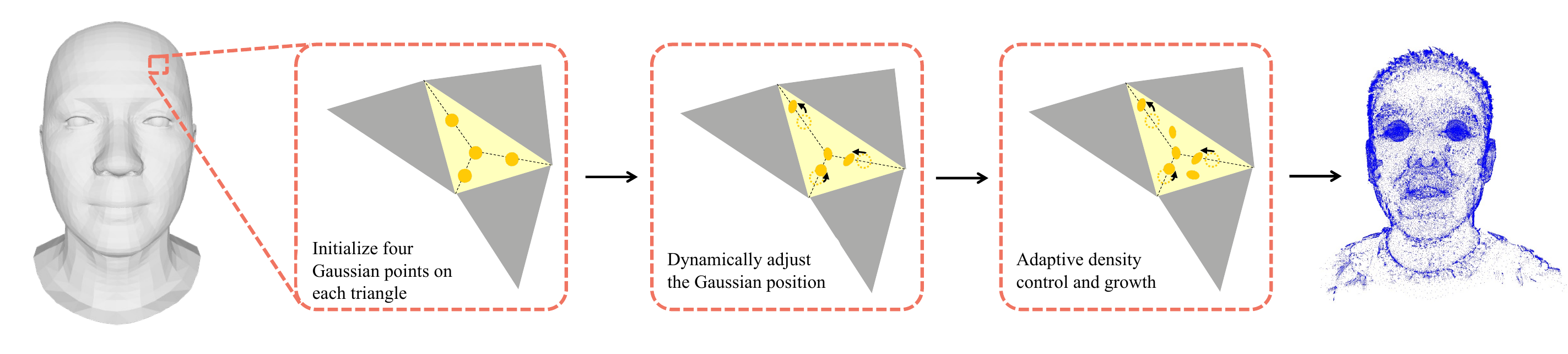}
\caption{
The pipeline of Point-based Learnable Representation Field initialization and growth. We allocate four Gaussian points of each triangle as initialization. During training, the positions of Gaussian points will be dynamically adjusted. Meanwhile, we adopt the adaptive density control and growth strategy, which adds and removes splats based on the viewspace positional gradient and the opacity of each Gaussian point.}
\label{fig:ex2}
\end{figure*}

\section{Method}
The proposed model is outlined in Fig. \ref{fig:ex1}. Firstly, we preprocess the given monocular video into FLAME parameters as detailed in Sec. \ref{sssec:num0}. During the initialization stage (Sec. \ref{sssec:num1}), we employ the Point-based Learnable Representation Field aligned with the canonical mesh, where the position dynamically adjusts throughout training. Subsequently, we fit the deformation between the canonical space and the current frame utilizing the transform network (Sec. \ref{sssec:num2}). To further enhance the geometry reconstruction performance, we introduce alpha rendering (Sec. \ref{sssec:num3}).

\subsection{Data Preprocessing}
\label{sssec:num0}
Given a monocular video $V$ consisting of images $I = \{I_i\}$, our objective is to extract the camera parameters including the intrinsic parameters $K_i$, the camera poses $C_i$, the FLAME meshes $M_i$ and corresponding properties set $\rho_i$ like shape, expression ,and the pose like jaw, neck, and eyes. 
Given that a monocular video provides only a single camera view, it inherently contains less information than the datasets used in Nersemble \cite{kirschstein2023nersemble}. This limitation necessitates more rigorous data preprocessing. In our experiments, the failure to align the initialized parametric mesh with the ground truth image results in disorganized results.

To adapt effectively to the monocular video setting, we fix the neck pose during processing and solely optimize the camera pose relative to the head, diverging from the approach in GaussianAvatars \cite{qian2023gaussianavatars}. Moreover, we replace the screen coordinates to Normalized Device Coordinates (NDC) using pytorch3D \cite{ravi2020pytorch3d}, ensuring greater compatibility with 3DGS. Our optimization focuses on the field of view (FOV) while keeping the properties $Z_{near}$ and $Z_{far}$. During preprocessing, we initially optimize the identity information within the FLAME2020 framework and subsequently optimize expressions and other properties while keeping the identity information fixed, effectively decoupling identity from expression information. Further details are available in the supplementary materials.

\subsection{Point-based Learnable Representation Field}
\label{sssec:num1}
Given the shape, pose and expression components, a morphologically realistic mesh can be produced via FLAME framework. The key is how to establish connection between the FLAME mesh with 3D Gaussian Splatting effectively. Motivated by GaussianAvatars \cite{qian2023gaussianavatars}, we build a Point-based Learnable Representation Field(PLRF) on the original meshes in order to augment the original initialized Gaussian point field. As shown in Fig. \ref{fig:ex2}, instead of initializing one Gaussian point at the center of each triangle face, we adopt the learnable strategy. Given the three vertices of a triangle set $\{x_1, x_2, x_3\}$, we first take their mean position $\bar x$ by:
\begin{equation}
    \bar x = mean(x_1, x_2, x_3)
\end{equation}

Via the barycentric coordinate $\bar x$, we get the lines with the three vertices of a triangle set $\{\bar x x_1, \bar x x_2, \bar x x_3\}$. Besides, we allocate the Gaussian points on the lines by:

\begin{equation}
\begin{aligned}
    x'_1 &= (1-n)*\bar x + n * x_1 \\
    x'_2 &= (1-n)*\bar x + n * x_2 \\
    x'_3 &= (1-n)*\bar x + n * x_3 
\end{aligned}
\end{equation}

where $n$ is a learned parameter ranging $[0, 1]$ with is initialized at 0.5. Thus, the new positions of the Gaussian point is represented as $\{x'_1, x'_2, x'_3, \bar x\}$. 

We attribute four Gaussian point at each triangle face when initialization, which is equivalent to splitting a complete triangle face into four sub-face. Additionally, we count the Gaussian points separately which are cloned during training. It is this approach named as PLRF that enhancing the density of the Gaussian points and avoiding the hole on the rendered avatars.

After reconstructing the point-based field, we adopt the global transform which transform the local Gaussian points from triangle face to the global space represented as:
\begin{align}
    r^{g} &= \mathbf{R}r \\
    \mu^{g} &= \frac{1}{2}\mathbf{k}\mathbf{R}\mu + \{x'_1, x'_2, x'_3, \bar x\} \\
    s^{g} &= \frac{1}{2}\mathbf{k}ns 
\end{align}

where $\{r, \mu, s\}$ are the Gaussian raw attributes in local space, $\{r^{g}, \mu^{g}, s^{g}\}$ are the Gaussian attributes in global space, $\mathbf{R} = \{R, R, R, R\}$, $\mathbf{k} = \{k, k, k, k\}$. We define $R$ as global rotation represent as the orientation of the triangle in the global space, $k$ by the mean length of one of the edges and its perpendicular as the triangle scaling. Here, we repeat the original scale and rotation four times to fit the alteration on positions. Each triangle owns its unique triangle scaling factor $k$ and global rotation $R$, which are fixed during training. 

Besides, we adopt the adaptive density control and growth. For each 3D Gaussian that exhibits a large view-space positional gradient, we either split it into two smaller Gaussian points if it is large, or clone it if it is small. 

Because of the learnable sampling strategy, to prevent the scale of a Gaussian point within a triangle face from being too large, we introduce the scale regularization on the scale prosperity, which we will describe in Sec. \ref{sssec:num4}.

\subsection{Transform Network}
\label{sssec:num2}
Here, we get the global Gaussian attribute in the global space $r^{g}$, $\mu^{g}$, $s^{g}$, which attach on mesh. However, there are still non-surface regions and subtle facial details can not be modeled by FLAME framework. To fit the Gaussian deformation due to the expression motions, we employ a standard MLP network $F_{\theta}$ for Gaussian deformation transform. Here, we define the Point-based Learnable Representation Field at timestep 0 as initial canonical space. The MLP network take the position of initial canonical space $\mu^{g}_0$ and current FLAME properties $\rho_i$ as expression and the pose except the shape property as input, and outputs spatial residuals of Gaussian attribute:
\begin{align}
    \delta \mu_i, \delta s_i, \delta r_i = F_{\theta}(\gamma(\mu^{g}_0), \rho_i)
\end{align}

where $\theta$ represents the optimized parameters of the standard Multi-Layer Perceptron (MLP), $\gamma$ represents the positional encoding of the spatial coordinates of the 3D Gaussian into a high-dimensional sequence as described in NeRF\cite{mildenhall2021nerf}. Here, we choose 10 as the frequency of the positional encoding. Final, the current spatial parameters which are input into the final render processing can be represented as:
\begin{align}
    \mu^{g}_i, s^{g}_i, r^{g}_i, = \mu^{g}_0 \oplus \delta \mu_i, s^{g}_0 \oplus \delta s_i, r^{g}_0 \oplus \delta r_i
\end{align}

\subsection{Geometry Enhancement}
\label{sssec:num3}
The hair strands or other model non-facial structures such as
eyeglasses and hairstyles make a profound impact on the photo-realistic rendering. In this work, we incorporate a alpha map $\hat{A}$ as each ground true view. We introduce the alpha rendering as follow:
\begin{equation}
    \hat{A}(x) = \sum_{i \in M}\alpha_{i}(x) \prod_{j=1}^{i-1} (1 - \alpha_{j}(x))
\end{equation}
We then encourage the consistency between the rendered alpha map $\hat{A}$ and the pseudo alpha map $A$, which is quantified as follows:
\begin{equation}
    \mathcal{L}_{\alpha} = \lambda_{\alpha}\| \hat{A}-A \|^2
\end{equation}
where $\lambda_{\alpha}$ is a hyperparameter equal to 0.5 and the pseudo $A$ is obtained from SGHM \cite{chen2022sghm}. 

\subsection{Optimization Scheme}
\label{sssec:num4}
Given the rendered image $\hat{C}$ via Eq. 3 and the ground truth image $C$, we utilize the L1 loss $\mathcal{L}_{l1}$ and a D-SSIM term \cite{zhang2018unreasonable} to supervise the pixel-wise difference by:
\begin{equation}
\begin{aligned}
    \mathcal{L}_{color} = (1-\lambda_{ssim})&\mathcal{L}_{l1} +\lambda_{ssim}\mathcal{L}_{D-SSIM} \\
    \mathcal{L}_{l1} &= \| \hat{C} - C \|^2  \\
\end{aligned}
\end{equation}
where $\lambda_{ssim}$ is a hyperparameter equal to 0.4. 

Besides, to maintain the structure information of the rendered image $\hat{C}$ with ground truth image $C$, we additionally add a structure loss $\mathcal L_{st}$ between them, to maintain the detail information and increase the contrast. 

\begin{equation}
    \mathcal L_{st} = \lambda_{st}((\nabla_{lr}(\hat{C})-\nabla_{lr}(C))^2 + (\nabla_{rl}(\hat{C})-\nabla_{lr}(C))^2),
\end{equation}
where $\nabla_{lr}$ denotes the gradients of image calculating from left to right, $\nabla_{rl}$ denotes the gradients of image calculating from right to left and $\lambda_{st}$ is a hyperparameter equal to 0.3. Here we manipulate a 2D convolution operation to calculate the gradients. Due to we expand the multi-view setting to single-view, there are some regularization items we introduce to make sure the consistency between the monocular video and the FLAME mesh. 

\textbf{Scale regularization on invisible point.} 
Under the multi-view setting, most Gaussian points around the original FLAME mesh are visible. However, since the constraints of monocular view, which captures only a limited number of points compared to multi-view perspectives, Gaussian points located in positions not visible from the monocular camera view may negatively impact rendering images under novel viewpoints.

Thus, we introduce the invisible point scale regularization by:
\begin{equation}
    \mathcal{L}_{invis} = \| \mathcal{M}*s \| 
\end{equation}
\begin{equation}
    \mathcal{M} = \begin{cases} 
        1, \mbox{if radii} < 0 \\
        0, \mbox{otherwise}
    \end{cases}
\end{equation}
where $\mathcal{M}$ represent the invisibility mask. 

\textbf{Scale threshold regularization.}
As for the visible Gaussian points, if the scale of
a Gaussian point within a triangle face is too large, it will result in unreasonable jittering artifacts. In order to mitigate this, we add the threshold regularization to the visible Gaussian points: 
\begin{equation}
    \mathcal{L}_{scale} = \left\| \max ((1 - \mathcal{M})s, \xi_{scaling} ) \right\|
\end{equation}
with $\xi_{scaling} = 0.15$. 

Our final loss function is thus:
\begin{equation}
    \mathcal{L} = \mathcal{L}_{color} + \mathcal{L}_{\alpha} + \mathcal L_{st} +\lambda_{invis} \mathcal{L}_{invis} + \lambda_{scale} \mathcal{L}_{scale}
\end{equation}
with $\lambda$ denoting the weights of each regularization term, which are set as follows: $\lambda_{invis} = 0.3, \lambda_{scale} = 0.15$.

\section{Experiments}
\subsection{Experiment Setup}

\begin{figure}[htb]
\centering
\includegraphics[width=8.3cm]{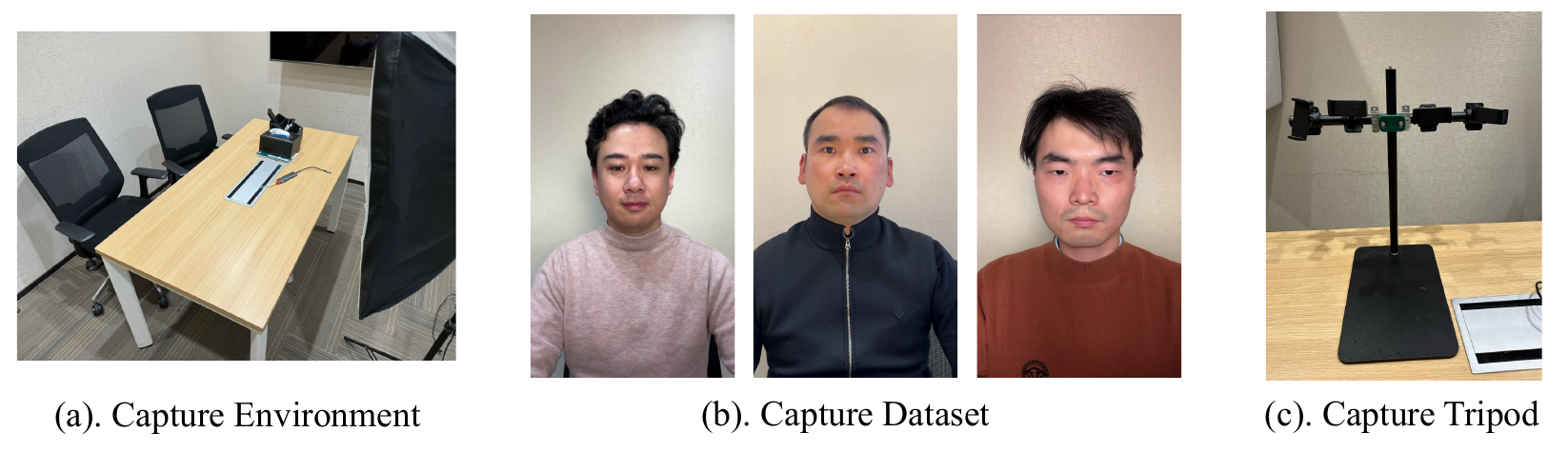}
\caption{
Capturing details of our dataset. }
\label{fig:ex11}
\end{figure}

\textbf{Dataset.}
In the experiment, we use 6 sets of data which mainly released by previous works \cite{xiang2023flashavatar, gao2022reconstructing}. Each dataset mainly contains a complete video with 25 FPS and 512*512 resolution. The length of the processed video is between 2 and 3 minutes, roughly 3000-4000 frames. We utilize RemBg for foreground segmentation and SGHM \cite{chen2022sghm} for alpha map. Here, all subjects are open-source. 

Besides, we capture our own dataset containing 3 subjects to verify the robustness of our approach using iPhone 15 Pro Max. We use a capture tripod to prevent the phone from shaking. The person sitting in front of the iPhone camera are prompted to rotate their heads and enact various face expressions, the capturing details are shown in Fig. \ref{fig:ex11}. All the captured videos will be cropped and down-sample to 600*600 resolution, each video contain approximately 1500-2000 frames. We use the preprocessing code to extract the Flame parameters. Further details of the data preprocessing pipeline are available in the supplementary material.

\textbf{Implementation Details.}
We implement our network with Pytorch \cite{paszke2019pytorch} and use Adam \cite{kingma2014adam} for parameter optimization. The learning rate of the Gaussian's parameters is the same of the original implementation, while the learning rate of the transform network is \(1e-4\). We train our model for 120000 iterations. Every 3000 iterations, we reset the opacity prosperity of Gaussian points. Then, we will perform a dynamic adaptation control strategy every 400 rounds until the 60000 iterations. We use a single Nvidia GeForce RTX4090 GPU for all of our experiments.

\textbf{Baseline.}
We compare our method with three representative works, as INSTA \cite{zielonka2023instant}, representing an efficient implicit head representation that creates a surface-embedded dynamic neural radiance field, based on neural graphics primitives; GaussianAvatars \cite{qian2023gaussianavatars}, a novel approach with Gaussian points bind to drive the Gaussian rendering; Flashavatar \cite{xiang2023flashavatar}, utilizing the UV sample as Gaussian initialization and offset network to eliminate the FLAME offset. For a fair comparison, both GaussianAvatars and FlashAvatar employ the FLAME2020 model, and their training settings are similar to ours. All experiments are conducted using a single Nvidia GeForce RTX 4090 GPU.

\begin{figure}
\centering
\includegraphics[width=6.cm]{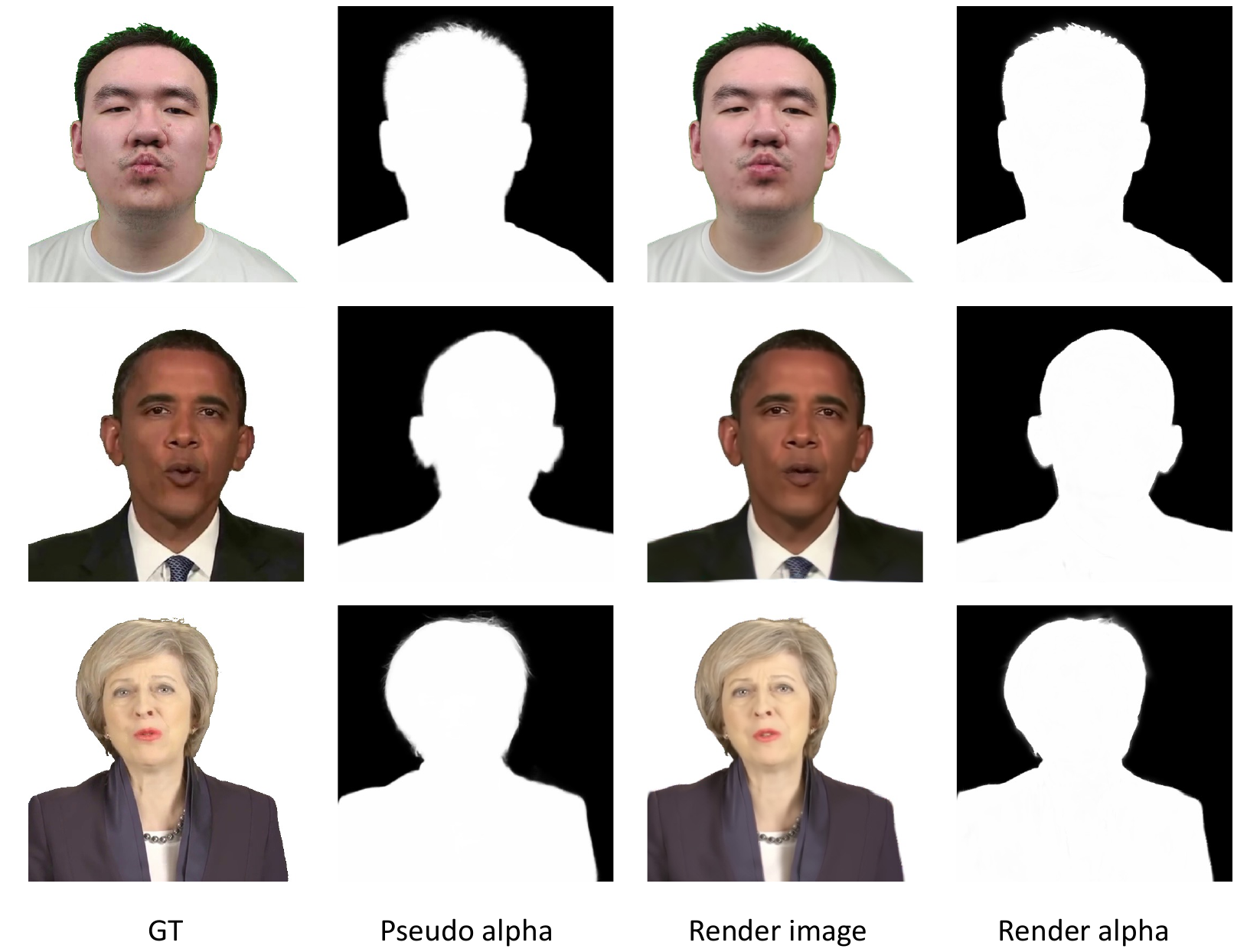}
\caption{
Alpha rendering results of FAGHead.}
\label{fig:ex3}
\end{figure}

\begin{figure*}
\centering
\includegraphics[width=10.cm]{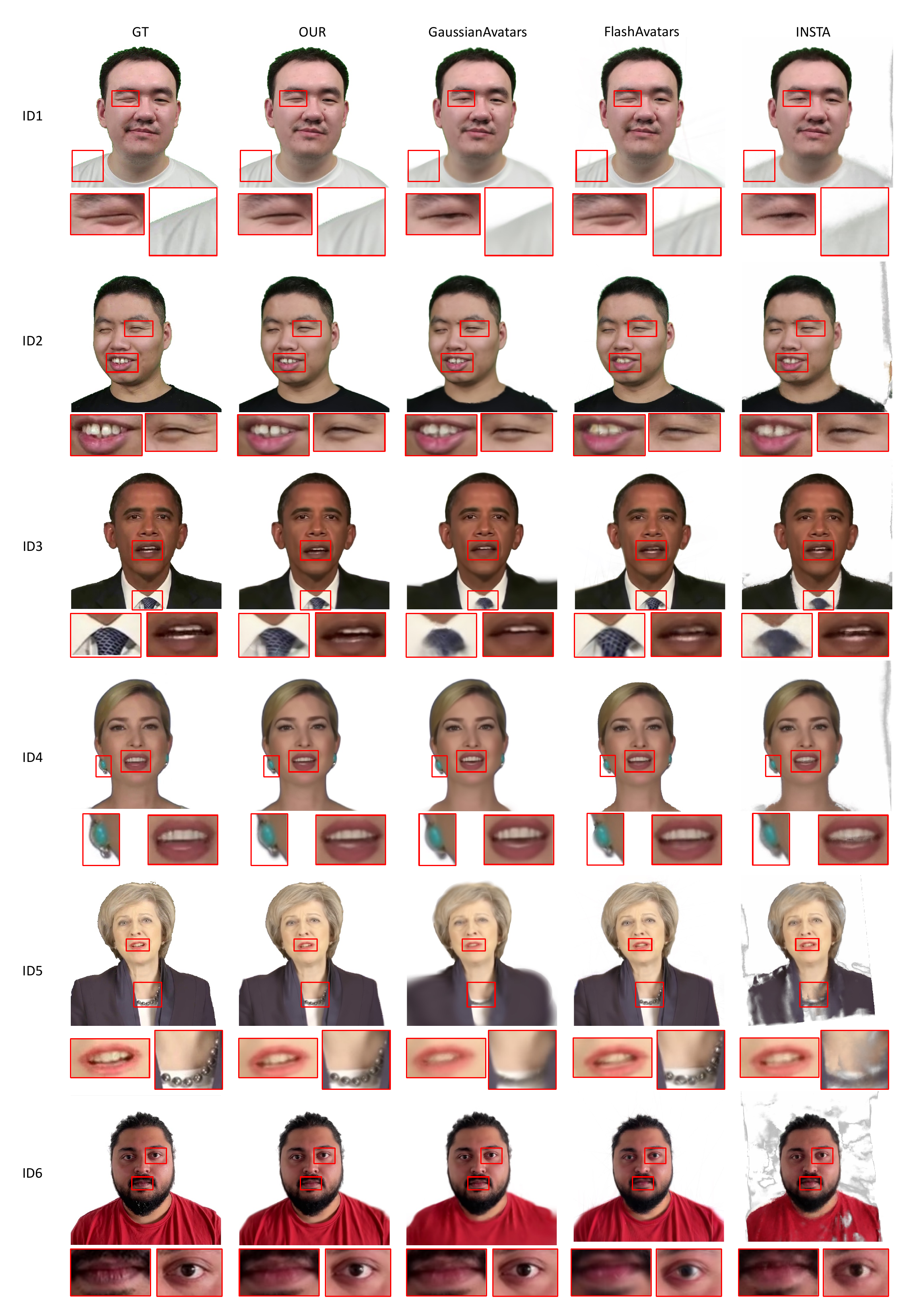}
\caption{
Qualitative comparison of ID 1-6 (from top to bottom) on the open-source datasets. FAGHead shows improved performances over strong baselines in capturing fine details such as shoulder strands, teeth, necklaces, etc..}
\label{fig:ex4}
\end{figure*}

\begin{figure*}
\centering
\includegraphics[width=10.cm]{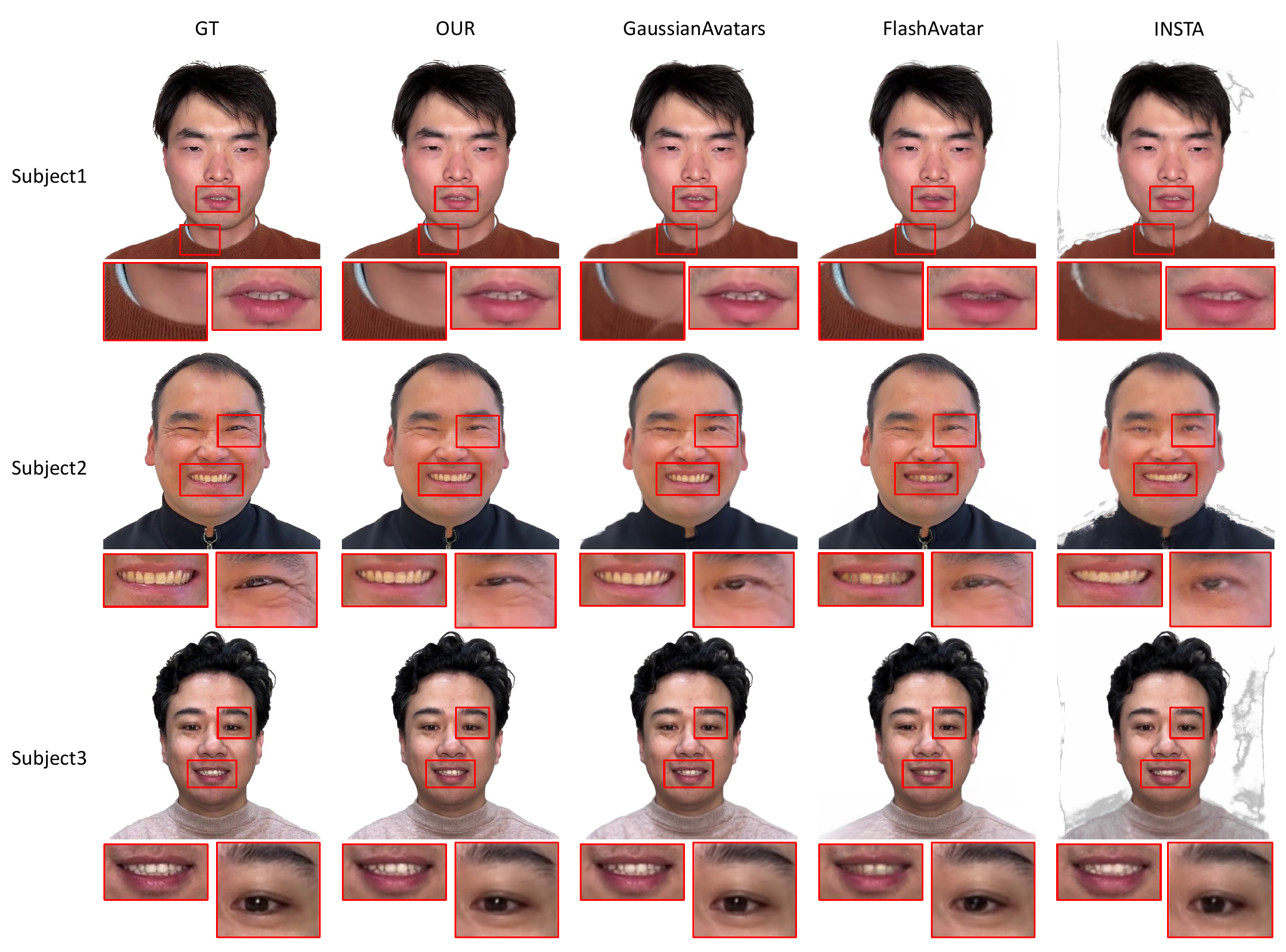}
\caption{
Qualitative comparison on subject 1-3 (from top to bottom) on our  capturing datasets. }
\label{fig:ex13}
\end{figure*}

\begin{figure*}[htb]
\centering
\includegraphics[width=11.cm]{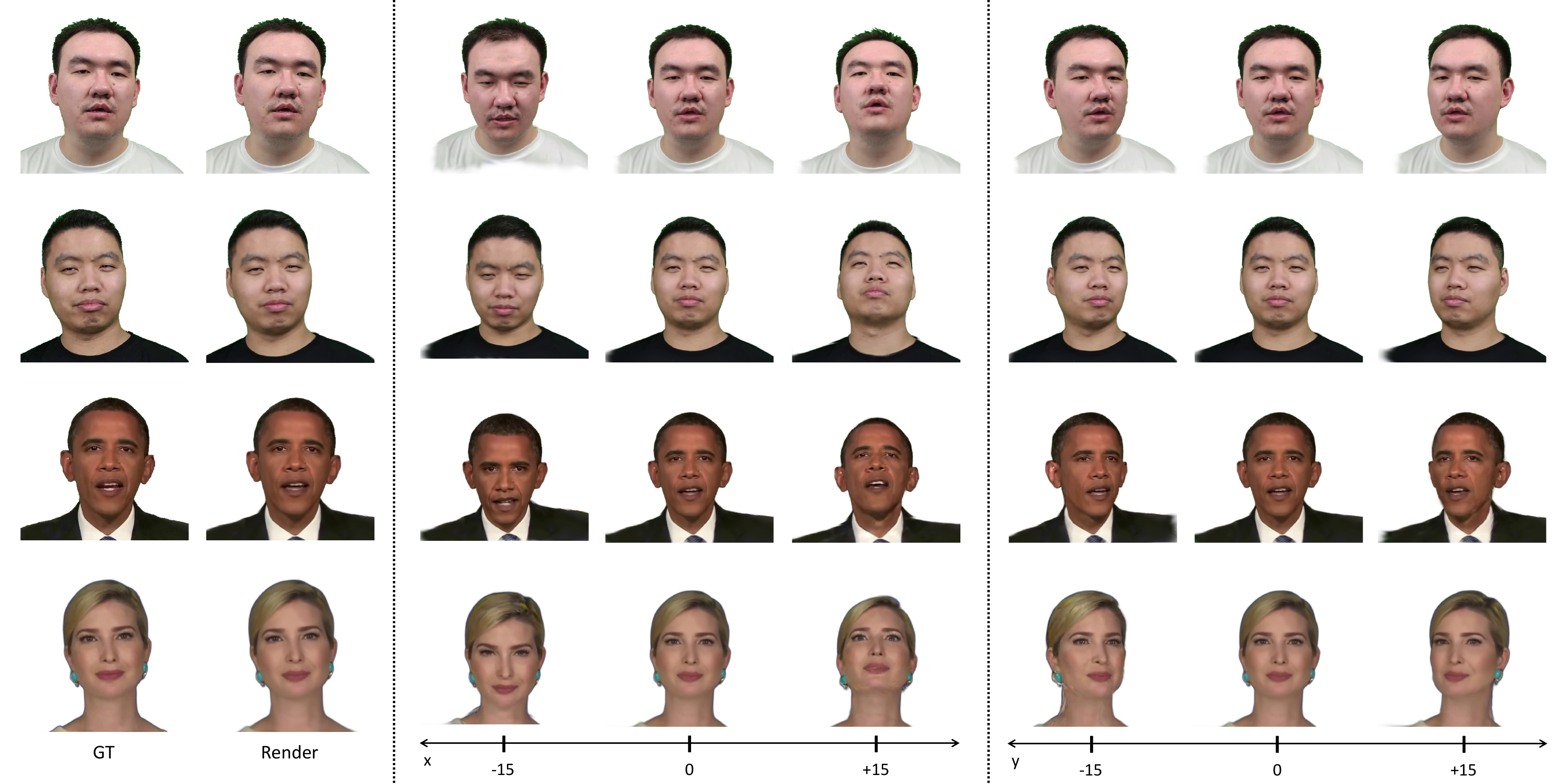}
\caption{
Novel view synthesis results of FAGHead. We demonstrate that it can  produce high-fidelity geometry and appearance from perspectives even not encountered during training.}
\label{fig:ex5}
\end{figure*}

\begin{table*}[h!]
    \centering
    \caption{Quantitative comparisons with state-of-the-art head avatar reconstruction method on the open-source datasets. }
    \resizebox{\textwidth}{!}{%
    \begin{tabular}{ccccccccccccccccccc}
    \toprule
    \textbf{Method} & \multicolumn{3}{c}{\textbf{"ID1"}} & \multicolumn{3}{c}{\textbf{"ID2"}} & \multicolumn{3}{c}{\textbf{"ID3"}} & \multicolumn{3}{c}{\textbf{"ID4"}} & \multicolumn{3}{c}{\textbf{"ID5"}} & \multicolumn{3}{c}{\textbf{"ID6"}}\\
    \cmidrule(lr){2-4} \cmidrule(lr){5-7} \cmidrule(lr){8-10} \cmidrule(lr){11-13} \cmidrule(lr){14-16} \cmidrule(lr){17-19} 
        & PSNR↑ & SSIM↑ & LPIPS↓ & PSNR↑ & SSIM↑ & LPIPS↓ & PSNR↑ & SSIM↑ & LPIPS↓ & PSNR↑ & SSIM↑ & LPIPS↓ & PSNR↑ & SSIM↑ & LPIPS↓ & PSNR↑ & SSIM↑ & LPIPS↓ \\
    \midrule
    INSTA \cite{zielonka2023instant}       
                & 17.28          & 0.852            & 0.207        
                & 15.95          & 0.831            & 0.221        
                & 14.63          & 0.846            & 0.177
                & 19.84	         & 0.921	        & 0.112
                & 13.83	         & 0.806	        & 0.251        
                & 12.56	         & 0.796	        & 0.272                           \\
    GaussianAvatars \cite{qian2023gaussianavatars}       
                & 25.12          & 0.906            & 0.156                         
                & 19.75          & \cellcolor{yellow!30}0.903            & 0.150                           
                & 17.77          & 0.907            & 0.138
                & 23.56	         & \cellcolor{yellow!30}0.952	        & \cellcolor{yellow!30}0.050
                & 19.02	         & 0.857	        & 0.222                               
                & 17.63	         & 0.782	        & 0.272                           \\
    FlashAvatar \cite{xiang2023flashavatar} 
                & \cellcolor{yellow!30}28.69 & \cellcolor{yellow!30}0.911   & \cellcolor{yellow!30}0.116                            
                & \cellcolor{yellow!30}25.46 & 0.902 & \cellcolor{yellow!30}0.105                           
                & \cellcolor{green!20}23.70  & \cellcolor{yellow!30}0.938   & \cellcolor{yellow!30}0.096
                & \cellcolor{yellow!30}25.05 & 0.951 & 0.075
                & \cellcolor{green!20}26.52  & \cellcolor{yellow!30}0.916	 & \cellcolor{yellow!30}0.097                               
                & \cellcolor{green!20}24.25  & \cellcolor{yellow!30}0.821   & \cellcolor{yellow!30}0.158                           \\
    Our         
                & \cellcolor{green!20}28.76  & \cellcolor{green!20}0.934    & \cellcolor{green!20}0.054                             
                & \cellcolor{green!20}25.81  & \cellcolor{green!20}0.935    & \cellcolor{green!20}0.062                           
                & \cellcolor{yellow!30}22.11 & \cellcolor{green!20}0.939    & \cellcolor{green!20}0.075
                & \cellcolor{green!20}26.92  & \cellcolor{green!20}0.963	 & \cellcolor{green!20}0.029
                & \cellcolor{yellow!30}25.14 & \cellcolor{green!20}0.918    & \cellcolor{green!20}0.074                           
                & \cellcolor{yellow!30}23.51 & \cellcolor{green!20}0.866	 & \cellcolor{green!20}0.128                           \\
    \hline
    \end{tabular}%
    }
    \label{tab:ex1}
\end{table*}

\begin{table*}[h]
    \centering
    \caption{Quantitative comparisons with state-of-the-art head avatar reconstruction method on our capturing datasets. }
    \resizebox{8cm}{!}{%
    \begin{tabular}{cccccccccc}
    \toprule
    \textbf{Method} & \multicolumn{3}{c}{\textbf{"subject1"}} & \multicolumn{3}{c}{\textbf{"subject2"}} & \multicolumn{3}{c}{\textbf{"subject3"}} \\
    \cmidrule(lr){2-4} \cmidrule(lr){5-7} \cmidrule(lr){8-10} 
        & PSNR↑ & SSIM↑ & LPIPS↓ & PSNR↑ & SSIM↑ & LPIPS↓ & PSNR↑ & SSIM↑ & LPIPS↓ \\
    \midrule
    INSTA \cite{zielonka2023instant}       
                & 17.38          & 0.812            & 0.242        
                & 18.07          & 0.835            & 0.236        
                & 15.31          & 0.771            & 0.264                         \\
    GaussianAvatars \cite{qian2023gaussianavatars}       
                & 23.02          & 0.859            & 0.144                         
                & 21.63          & 0.893            & 0.154                           
                & 26.49          & \cellcolor{yellow!30}0.904            & 0.088                        \\
    FlashAvatar \cite{xiang2023flashavatar} 
                & \cellcolor{yellow!30}26.68 & \cellcolor{yellow!30}0.860   & \cellcolor{yellow!30}0.079                             
                & \cellcolor{yellow!30}27.23 & \cellcolor{yellow!30}0.907   & \cellcolor{yellow!30}0.076                           
                & \cellcolor{yellow!30}28.48 & 0.900 & \cellcolor{yellow!30}0.074\\
    Our         
                & \cellcolor{green!20}27.83 & \cellcolor{green!20}0.900   & \cellcolor{green!20}0.066                             
                & \cellcolor{green!20}28.31 & \cellcolor{green!20}0.929   & \cellcolor{green!20}0.064                          
                & \cellcolor{green!20}30.61 & \cellcolor{green!20}0.941   & \cellcolor{green!20}0.041\\
    \hline
    \end{tabular}%
    }
    \label{tab:ex3}
\end{table*}

\subsection{Qualitative and Quantitative Comparison in Reconstruction}
Fig. \ref{fig:ex3} visualizes the alpha rendering result of FAGhead. 
Fig. \ref{fig:ex4} and Fig. \ref{fig:ex13} depict the qualitative comparison between our model and the above method.

INSTA \cite{zielonka2023instant} is able to generate renders that are reasonably consistent with articulated facial expression and head-pose. However, INSTA utilizes neural graphics primitives integrated within the FLAME surface, which limits its ability to accurately model accessories such as ties and necklaces(in the 3rd and 5th rows). Besides, it tends to generate artifacts around the mouth(the 2nd and 5th row in Fig. \ref{fig:ex4} and the 1st row in Fig. \ref{fig:ex13}) and smooth results with ignoring thin structures, especially in the shoulder region. 

GaussianAvatars \cite{qian2023gaussianavatars} primarily relies on a parametric morphable face model to rig 3D Gaussian splats. It achieves a more precise geometric representation by attaching the 3D Gaussian splats from local to global coordinates within a FLAME mesh, indicating that the rendering performance hinges on the quality of the parametric morphable face model. Thus, it tend to generate artifacts around eyes because of the geometry errors(the 1st row in Fig. \ref{fig:ex4}). Besides, due to exploitation of the parametric morphable face model, it usually leads to a over-smooth results of the accessories(the 3rd and 5th rows in Fig. \ref{fig:ex4} and the 1st row in Fig. \ref{fig:ex13}).

FlashAvatar \cite{xiang2023flashavatar} employs a uniform 3D Gaussian field embedded in the surface to initialize. However, becasue of the fix number of 3D Gaussian splats without the adaptive density control strategy, artifacts will show up as spikes at the edge of the avatars like shoulder or hair(in the 1st rows). Moreover, blurs will appear around the eye region and mouth region(the 2nd and 3rd row in Fig. \ref{fig:ex4} and 3rd row in Fig. \ref{fig:ex13}). In contrast, our method produces photorealistic images that closely align with the ground truth, capturing nearly all fine facial details, thin structures, and accessories.

Please see the Tab. \ref{tab:ex1} and Tab. \ref{tab:ex3} for the quantitative comparison between our model and other method. The metrics include PSNR, SSIM, and LPIPS \cite{zhang2018unreasonable}.

\begin{figure}[htb]
\centering
\includegraphics[width=6cm]{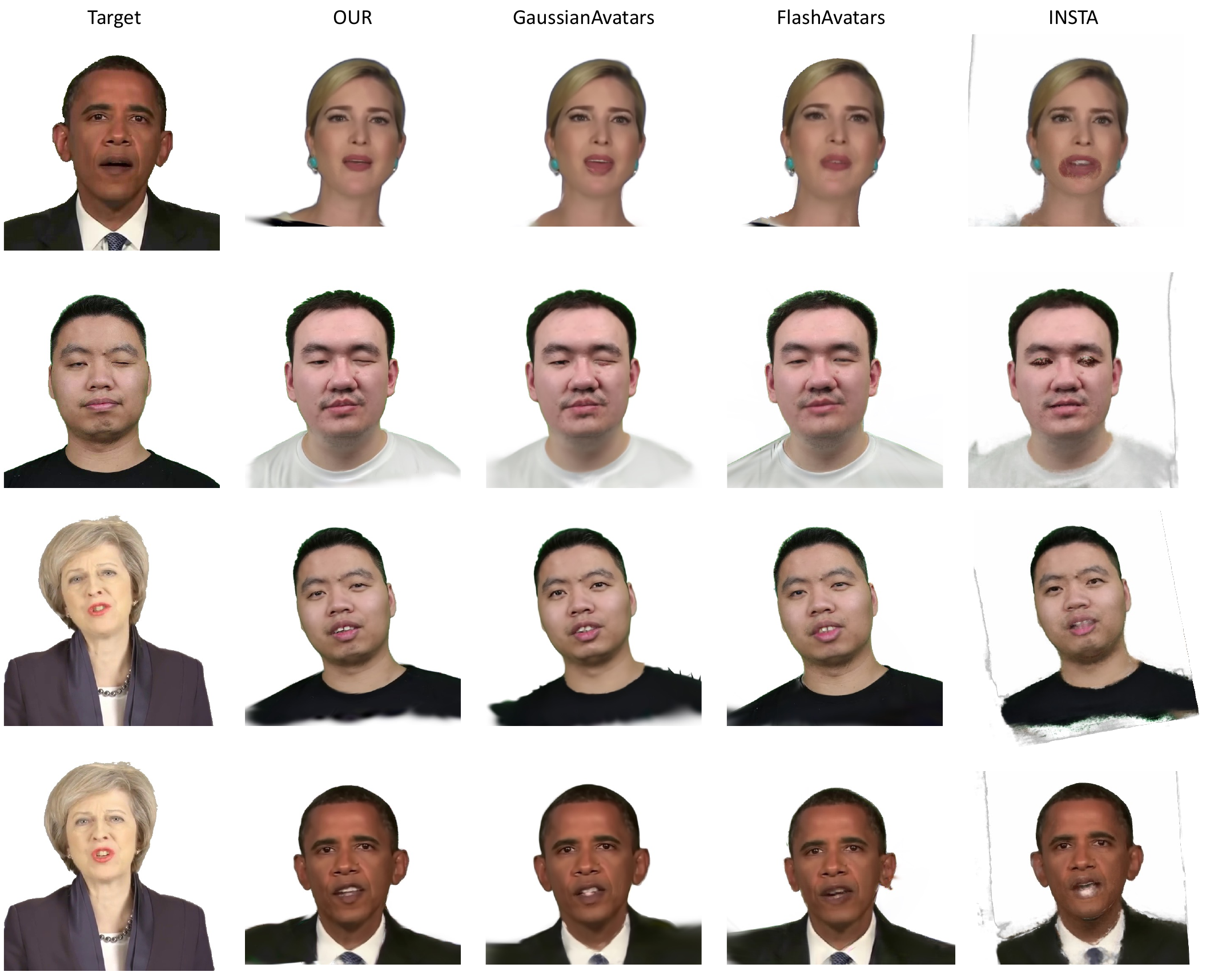}
\caption{
Qualitative results of ours and three other method on
facial reenactment task. Our method preserves personalized facial
details in gaze direction, and interior mouth regions and synthesizes
more natural results.}
\label{fig:ex6}
\end{figure}

\begin{figure}[htb]
\centering
\includegraphics[width=5.5cm]{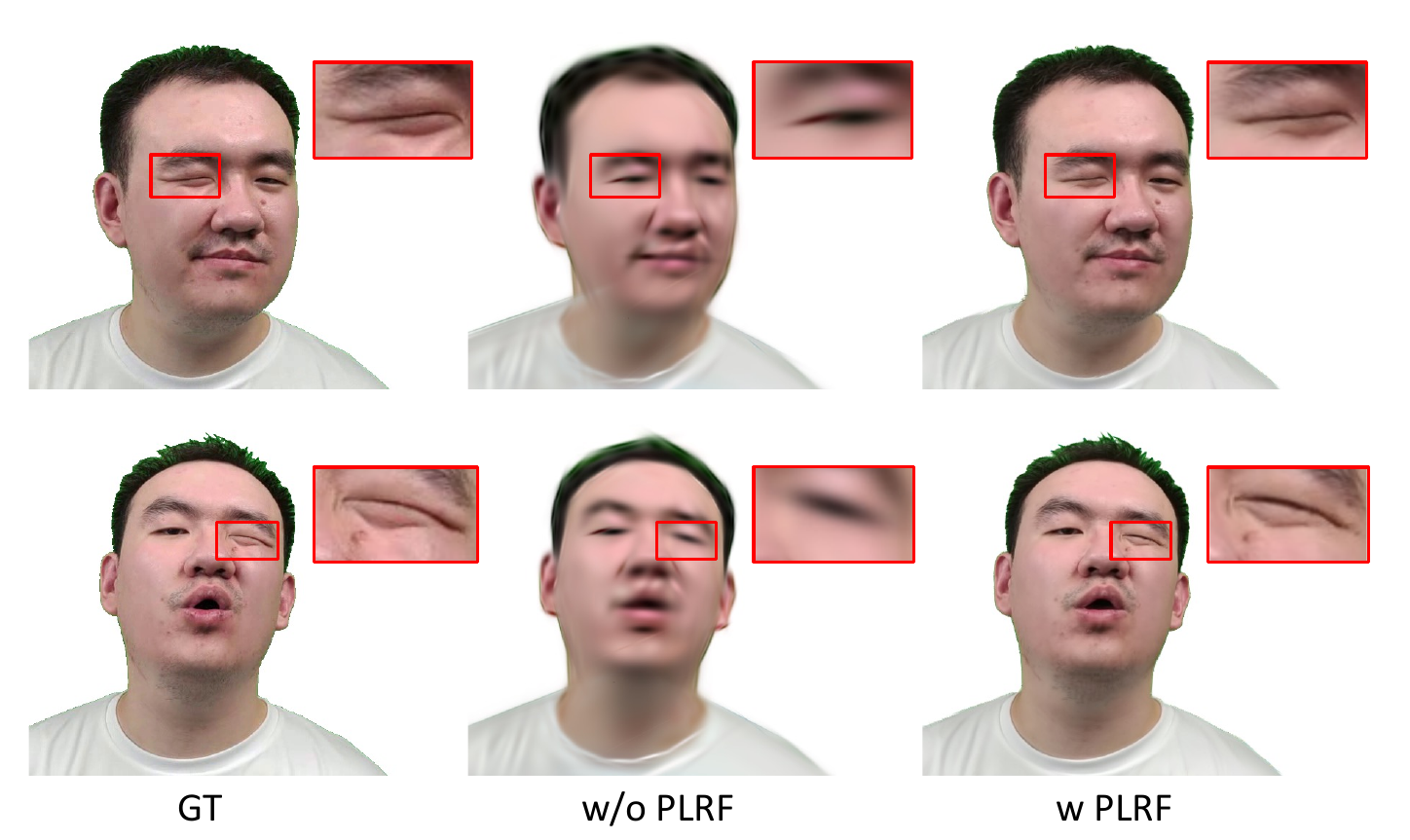}
\caption{
The effect of Point-based Learnable Representation.}
\label{fig:ex7}
\end{figure}

\begin{figure}[htb]
\centering
\includegraphics[width=5.5cm]{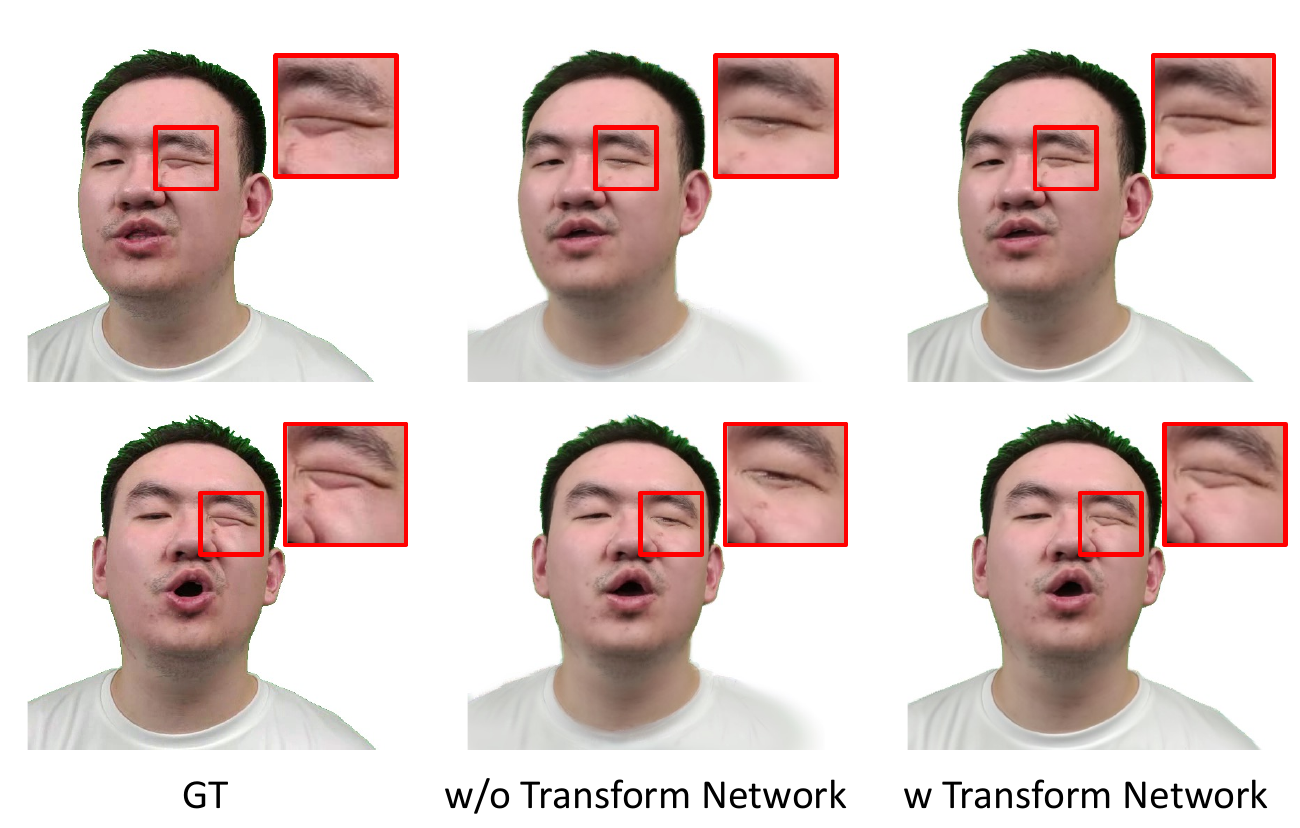}
\caption{
The effect of Transform network.}
\label{fig:ex8}
\end{figure}
\vspace{-0.3cm}
\begin{figure*}[htb]
\centering
\includegraphics[width=9.5cm]{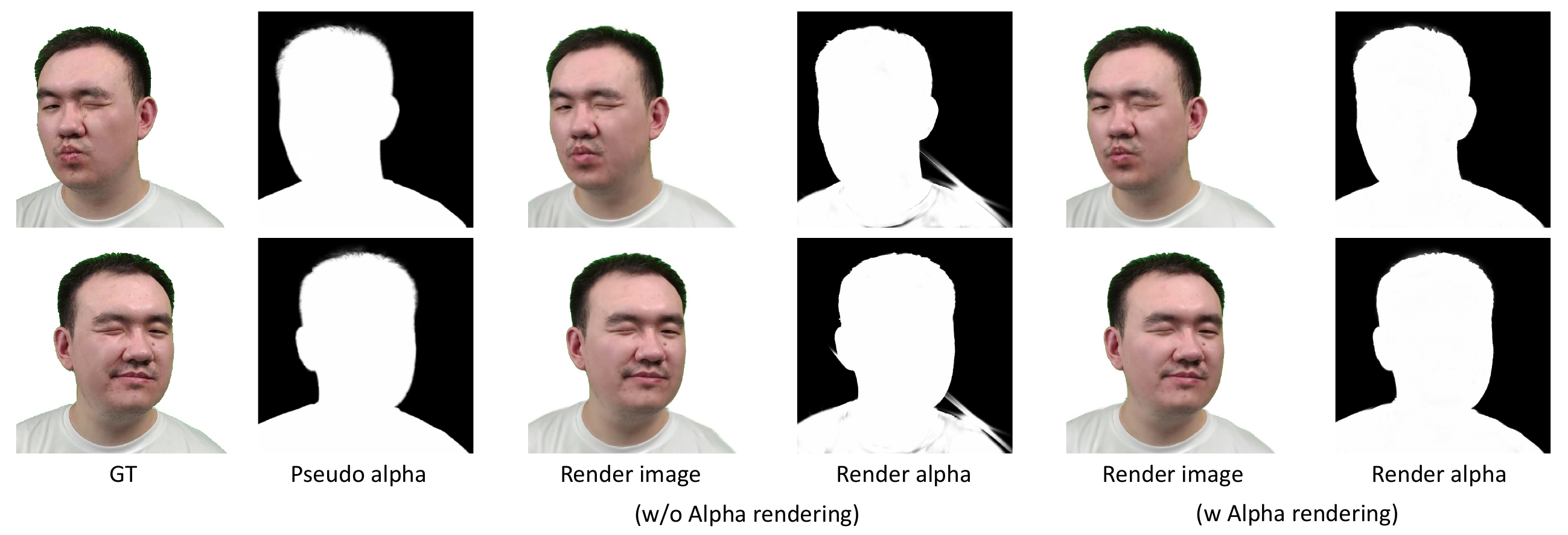}
\caption{
The effect of Alpha rendering.}
\label{fig:ex9}
\end{figure*}
\vspace{-0.3cm}

\subsection{Novel View Synthesis}
To demonstrate the novel view rendering ability of our FAGhead, we freely rotate the camera pose to generate new results from novel rendering views as illustrated in Fig. \ref{fig:ex5}. We first utilize a specified sample from the test dataset with FLAME parameter and camera pose. The multiple viewpoints can be produced via multiplying the camera pose with a novel rotation angle. Then the rendered images with different rendering views are generated by the Gaussian rendering. The novel view results show no artifacts or unrealistic facial expressions, even at intermediate rotation angles.

\subsection{Cross-identity Reenactment}
The cross-identity reenactment results are presented in Fig. \ref{fig:ex6}. Via replacing the input FLAME parameters of transform network, our approach can conduct the expression reenactment with finer-grained teeth and gaze direction. 

Compared with the previous works, GaussianAvatars \cite{qian2023gaussianavatars} shows the artifacts in shoulder edge and the unnatural gaze direction. With depending on the offset network, the results generated by FlashAvatar \cite{xiang2023flashavatar} seems a bit over-smooth. As for INSTA \cite{zielonka2023instant}, despite its deformable neural radiance field, tends to lose details in geometry. More reenactment results are available in the supplementary materials. 

\begin{table}[h]
    \centering
    \caption{We systematically ablated several key components and assessed quantitative performance to demonstrate their effectiveness. }
    \resizebox{6cm}{!}{%
    \begin{tabular}{lccc}
       \toprule
       \begin{tabular}[c]{@{}c@{}}\end{tabular} Metrics & PSNR↑ & SSIM↑ & LPIPS↓ \\
       \midrule
       w/o PLRF & 25.61 & 0.884 & 0.239 \\
       w/o Transform Network & 25.21 & 0.902 & 0.104 \\
       w/o Alpha rendering & 28.11 & 0.927 & 0.065 \\
       w/o Construct loss $\mathcal L_{st}$ & \cellcolor{yellow!30}28.47 & \cellcolor{yellow!30}0.931 & \cellcolor{yellow!30}0.058 \\
       Our & \cellcolor{green!20}28.75 & \cellcolor{green!25}0.934   & \cellcolor{green!25}0.054  \\
       \bottomrule
    \end{tabular}
    }
    \label{tab:ex2}
\end{table}

\subsection{Ablation Study}
To validate the effectiveness of our method components, we deactivate
each of them and report results in Tab. \ref{tab:ex2}. 

\textbf{Point-based Learnable Representation Field.} 
Without Point-based Learnable Representation Field, we randomly initialize 100000 Gaussian points as for creating Gaussian splatting as canonical space. With directly optimizing the Gaussian point attributes, we obtain the final results with spatial residuals output from the transform network. As shown in the 1st row of Tab. \ref{tab:ex2} and Fig. \ref{fig:ex7}, the expression details are not well captured,  demonstrating the influence of Point-based Learnable Representation Field in preserving fine details.

\textbf{Transform Network.}
Without transform network, we just utilize the raw FLAME parameters to fit the transform mesh rely on the linear transformations defined by the LBS formula from the original FLAME model\cite{li2017learning}. With the limited geometry information, it causes blurry renderings, leading to huge fidelity loss, as shown in the 2nd row of Tab. \ref{tab:ex2} and Fig. \ref{fig:ex8}.

\textbf{Alpha rendering.}
Without alpha rendering, artifacts will show up as spikes at the edge of the avatars like shoulder or hair. Besides, the hole will appear between the neck and the collar as shown in the 3rd row of Tab. \ref{tab:ex2} and Fig. \ref{fig:ex9}, which is absolutely unreasonable. Because the alpha rendering make sure the geometry consistency, without it, the Gaussian rendering process relies solely on color information and disregards geometric attributes.

\textbf{Construct loss $\mathcal L_{st}$.}
Construct loss $\mathcal L_{st}$ maintain the detail information and increase the contrast. We shown the results without the construct loss $\mathcal L_{st}$ in the 4th row of Tab. \ref{tab:ex2}.

\section{Conclusion and Discussion}
FAGhead is a novel approach which achieve high-fidelity reconstruction and full animation of 3D human avatars. We propose the Point-based Learnable Representation Field as the prior to reconstruct portraits and exploit the transform network to fit the deformation, outperforming state-of-the-art method. However, our method still has room for further improvement. One limitation is that it does not model the oral cavity effectively. Moreover, our rendering performance is heavily dependent on the quality of data preprocessing, indicating that we struggle to handle significant errors in this stage effectively. Addressing these issues will be a key focus of our future research efforts.



\clearpage  

%
%
\bibliographystyle{splncs04}
\bibliography{main}

\clearpage

\title{Supplementary Material}
\section{Implementation Details}
\subsection{Network Architecture}
\begin{figure}
\centering
\includegraphics[width = 9cm]{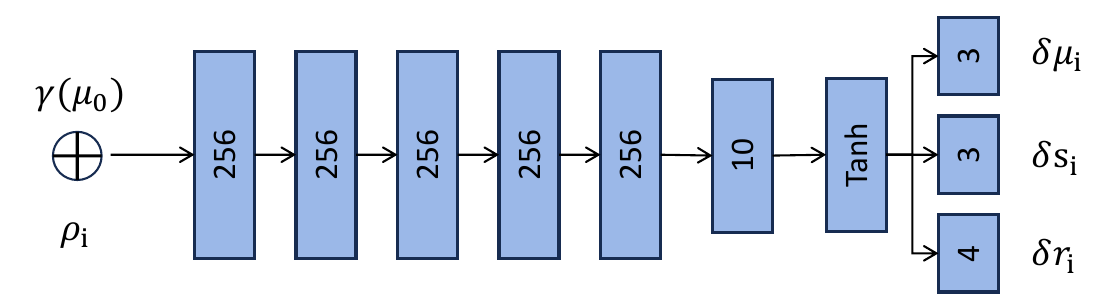}
\caption{
Network architecture of the transform network. }
\label{fig:ex15}
\end{figure}

In Fig. \ref{fig:ex15}, we show the transform network structure, we mainly utilize a set of Fully Connected(FC) layers with Tanh activation function. 

\subsection{Datapreprocess Pipeline}
\begin{figure}
\includegraphics[width = 9.5cm]{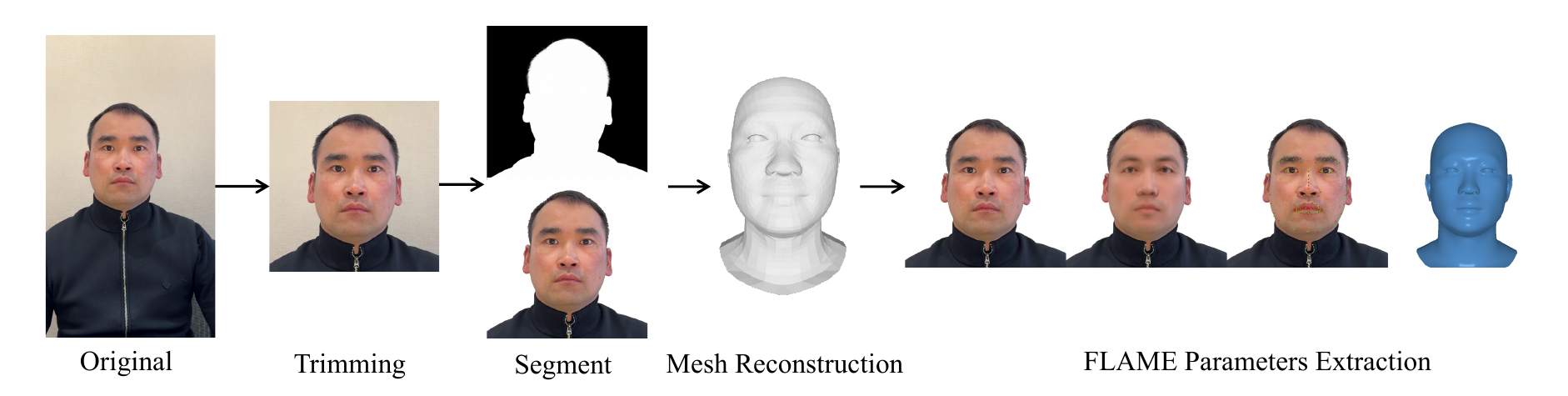}
\caption{
Data preprocessing pipeline. }
\label{fig:ex16}
\end{figure}

The data preprocessing pipeline is illustrated in Fig. \ref{fig:ex16}. After obtaining the original capture videos, we first crop the head region. The background is then removed using a segmentation algorithm, and the shape vectors of the mesh are extracted via the MICA. Finally, the FLAME parameters are extracted using our modified tracker based on the Metrical Photometric Tracker.

Here, we conduct an ablation study as shown in Fig. \ref{fig:ex17}, where we utilize the original tracker to extract the FLAME parameters. However, this leads to misalignment of the initialized mesh with the ground truth image, resulting in disorganized results.

\begin{figure}
\centering
\includegraphics[width = 5.5cm]{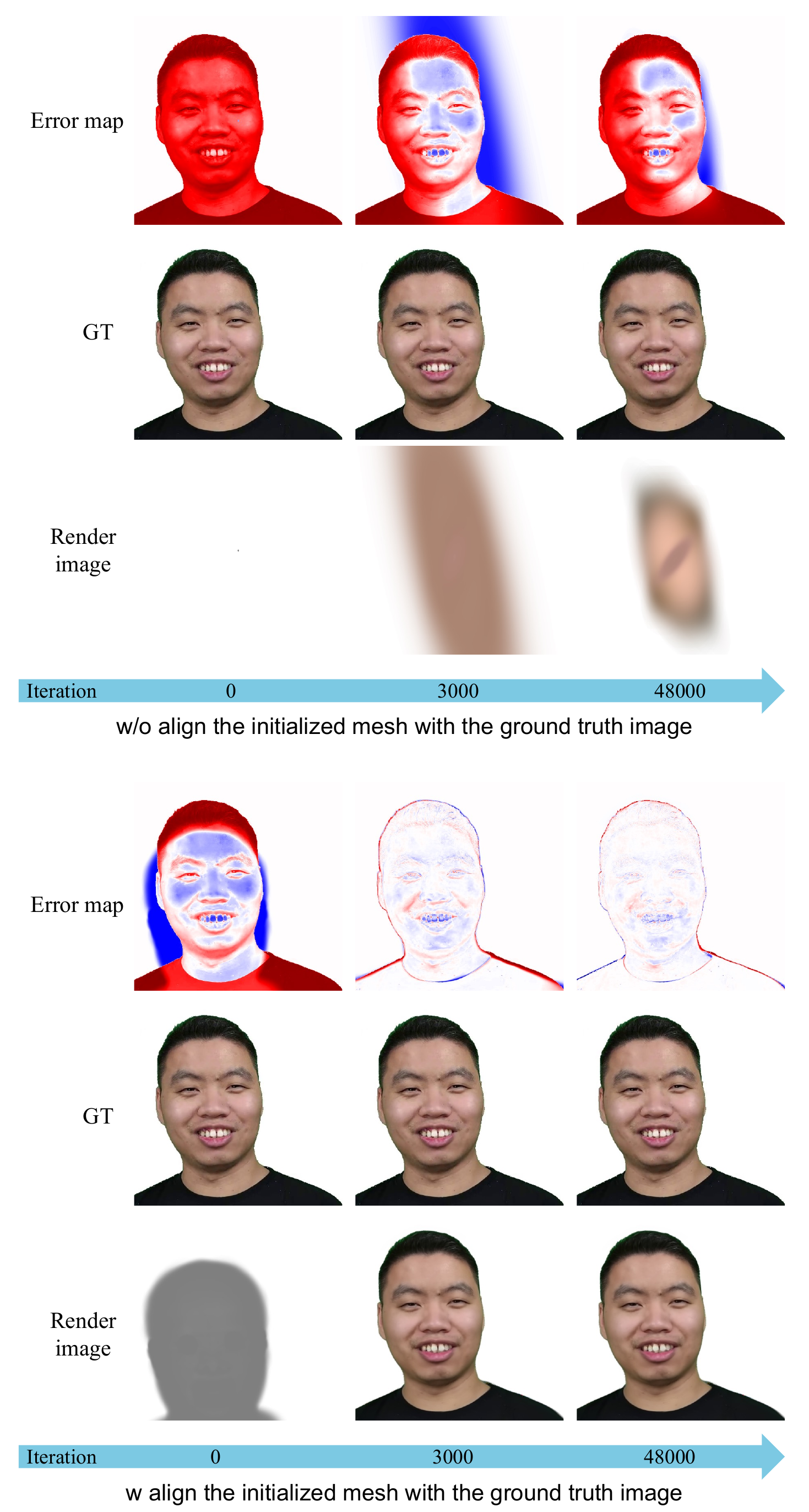}
\caption{
The effect of alignment the initialized mesh with the ground truth image. }
\label{fig:ex17}
\end{figure}

\section{Additional Results}
\subsection{Additional Reenactment Results}
In Fig. \ref{fig:ex17}, we present the additional reenactment results on our capturing datasets. Our approach demonstrates superior performance compared to the previous works. 

\begin{figure}
\centering
\includegraphics[width = 6.5cm]{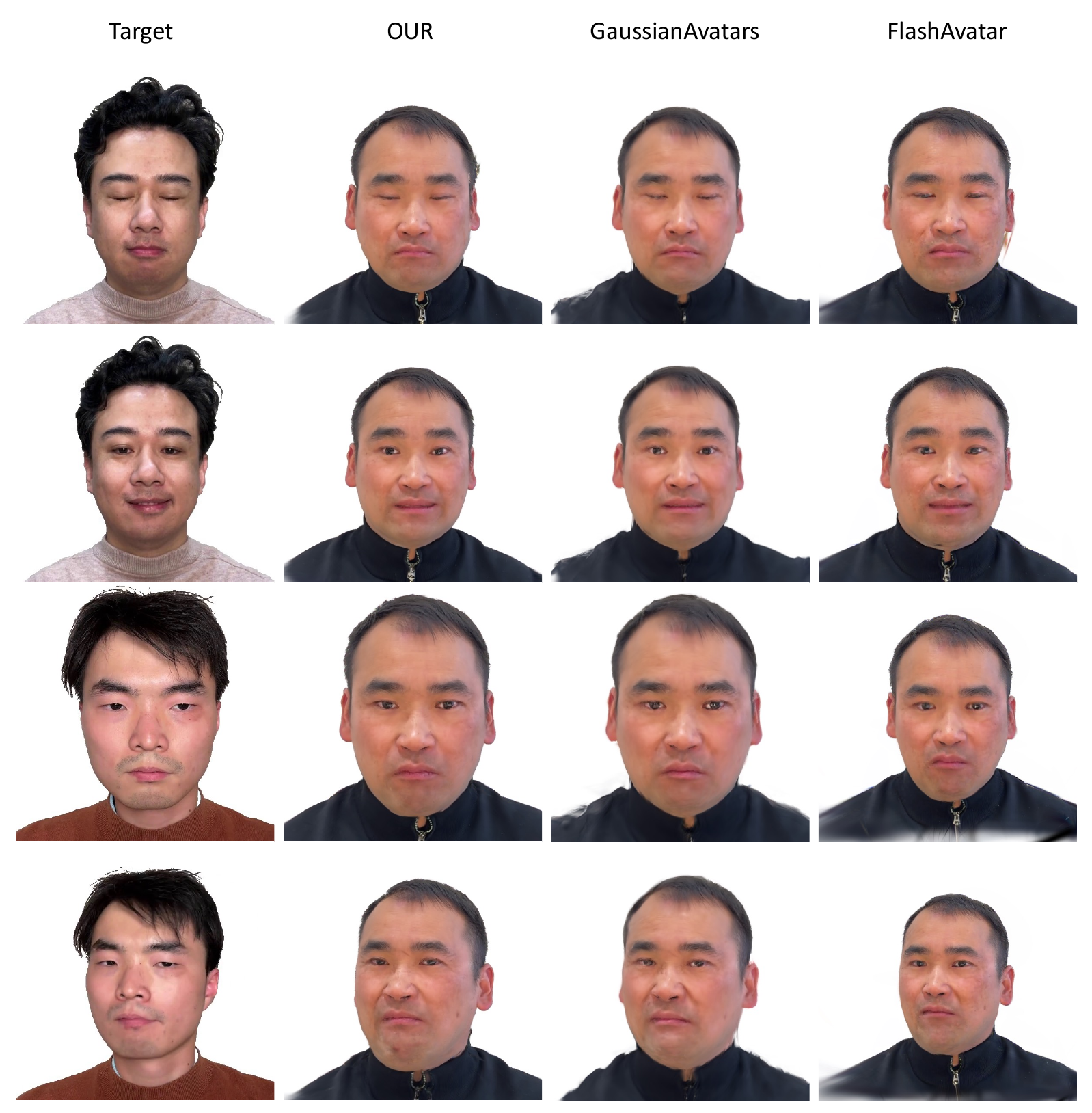}
\caption{
More qualitative results of ours and two other method on
facial reenactment task. }
\label{fig:ex17}
\end{figure}
\end{document}